\documentclass{article}
\usepackage{ijcai21}

\usepackage{times}
\usepackage{soul}
\usepackage{url}
\usepackage[hidelinks]{hyperref}
\usepackage[utf8]{inputenc}
\usepackage[small]{caption}
\usepackage{graphicx}
\usepackage{amsmath}
\usepackage{amsthm}
\usepackage{booktabs}
\usepackage{multirow}
\usepackage{amssymb}

\usepackage{algorithm}
\usepackage{algpseudocode}

\title{Channel DropBlock: An Improved Regularization Method for Fine-Grained Visual Classification
}

\author{
Yifeng Ding$^1$, Shuwei Dong$^1$, Yujun Tong$^1$, Zhanyu Ma$^1$, Bo Xiao$^1$, and Haibin Ling$^2$
\affiliations
$^1$Pattern Recognition and Intelligent System Laboratory, Beijing University of Posts and Telecommunications\\
$^2$Department of Computer Science, Stony Brook University
}

\begin{document}

\maketitle

\begin{abstract}
  Classifying the sub-categories of an object from the same super-category (\emph{e.g.}, \textit{bird}) in a fine-grained visual classification (FGVC) task highly relies on mining multiple discriminative features. Existing approaches mainly tackle this problem by introducing attention mechanisms to locate the discriminative parts or feature encoding approaches to extract the highly parameterized features in a weakly-supervised fashion. In this work, we propose a lightweight yet effective regularization method named Channel DropBlock (CDB), in combination with two alternative correlation metrics, to address this problem. The key idea is to randomly mask out a group of correlated channels during training to destruct features from co-adaptations and thus enhance feature representations. Extensive experiments on three benchmark FGVC datasets show that CDB effectively improves the performance.
\end{abstract}

\section{Introduction}
\label{sec:Introduction}
Deep convolutional neural networks have achieved great progresses in many computer vision tasks, especially in object recognition with sophisticated model design and abundant data sets. By contrast, Fine-Grained Visual Classification (FGVC) remains very challenging, mainly because of the high intra-class variance and low inter-class variance among fine-grained categories (e.g., \textit{bird species}, \textit{flower types}, \textit{car models}, \textit{etc}.).

The key in FGVC is to learn discriminative feature representations that can distinguish the inter-class differences and align the intra-class variances. Early solutions~\cite{chai2013symbiotic,zhang2014part} utilize additional bounding box/part annotations to locate discriminative parts. It later becomes clear that such supervised approaches are not optimal because expert human annotations can be cumbersome to obtain and often error-prone~\cite{zheng2017learning}. More recent methods~\cite{lin2015bilinear,zheng2017learning,fu2017look,yang2018learning} address this issue in a weakly supervised manner. However, these methods usually involve complicated network structure or highly parameterized feature representations, which is hardly applicable to other models and also introduces extra computation overhead in both training and inference stages.

To address the aforementioned concerns, in this paper we propose a novel lightweight yet powerful method, named Channel Dropblock (CDB), for weakly supervised FGVC. Since the features in the convolutional layers are correlated, CDB aims at eliminating the effect of co-adaptations among channels. This encourages the network to enhance feature representations and to find more discriminative visual evidence in a FGVC task. To this end, CDB is designed to drop a group of correlated channels in one or several of convolutional layers to motivate the network to distinguish more discriminative parts. Unlike most existing weakly supervised FGVC networks, CDB is a light structure without additional parameters, making it easily to integrate into existing networks.

To guide the channel selection, we propose two novel strategies to measure the channel correlations. One is the max activation metric, inspired by MA-CNN~\cite{zheng2017learning}, that measures channel correlation by computing the distance between peak responses from different channels. We suppose that the smaller the distance is, the closer correlation between two channels is. In this occasion, the channels are clustered into discriminative local regions with diverse activation centers. The other one is bilinear pooling metric. It computes the channel correlation matrix in which the cosine similarities between channels are calculated pairwisely. In the bilinear pooling metric, the larger the cosine similarity value is, the more similar the two channels are. With this metric, the channels are clustered into specific visual patterns. During training, one specific channel of the feature map is select randomly, and then the correlated channels which share close visual pattern are clustered and masked out based on the correlation matrix. 

Compared with existing FGVC networks, the proposed method is more efficient with flexibility. This is because it can erase a discriminative visual pattern to prevent the features from co-adaptations by a single forward-backward propagation in a single block, which can be easily applied to all kinds of networks. Moreover, it does not need extra part/object annotation and introduces no computational overhead at inference time.

Our contributions can be summarized as follows:

1) We address the challenges of discriminative feature learning in FGVC tasks by proposing a novel lightweight regularization structure, which drops a group of correlated channels to motivate the network enhancing feature representations and hence extract more discriminative patterns.

2) We propose two metrics to measure the pairwise correlation between different feature channels, which can draw insights into feature channels.

3) We conduct extensive experiments on three popular fine-grained benchmark datasets, the results demonstrate that the proposed CDB significantly improves the FGVC performance when applied to baseline networks or integrated into existing methods.

\section{Related Work}
\label{sec:RelatedWork}
\subsection{Dropout Mechanism}



Dropping some units or patches in CNN has been proposed as a regularization method, such as dropout~\cite{srivastava2014dropout}, SpatialDropout~\cite{tompson2015efficient}, cutout~\cite{devries2017improved} and DropBlock~\cite{ghiasi2018dropblock}, in which dropout is the inspiration of most other related regulation methods. Based on the way to drop, these methods can be divided into two categories: one contains attention-based methods and the other one drops object information randomly.

Dropout~\cite{srivastava2014dropout} is set to omit each neuron with a probability $p$ during training time, providing a simple and effective way to avoid overfitting. However, features are correlated spatially that hinders dropout to work effectively when applied to convolutional layers. To solve the problem, SpatialDropout~\cite{tompson2015efficient}, cutout~\cite{devries2017improved} and DropBlock were proposed later. Instead of dropping several inconsecutive units in feature maps, in SpatialDropout~\cite{tompson2015efficient}, entire feature maps are dropped with a probability $p$ randomly, preventing nearby pixels from presenting the same information as dropped neurons. Cutout~\cite{devries2017improved} is applied on input images to erase one part with a random square mask before training. Similarly, the key of DropBlock~\cite{ghiasi2018dropblock} is set to drop contiguous regions of feature maps, declining the effect of spatial correlated features. ADL~\cite{choe2019attention} is one of attention based regulation methods, in which drop masks based on attention are applied to feature maps to hide most discriminative parts and activate networks to learn other important features.

Different from dropping channels randomly of SpatialDropout~\cite{tompson2015efficient} and erasing a patch in input image randomly of Cutout~\cite{devries2017improved}, our method is set to drop a group of related channels. In this way, our method reduces the effect of co-adaptations among different channels, and thus can locate more discriminative visual patterns and enhance the features representation in FGVC.

\subsection{Fine-grained Classification}

The key of fine-grained image classification is to find out the discriminative representations of each class. To this end, a variety of methods have been proposed to increase the accuracy of deep learning network for this task. 

Bilinear CNN~\cite{lin2015bilinear} consists of two extractors, capturing pairwise correlations between the feature channels, to distinguish images from subtle differences, which is an efficient way to obtain more representation of features. RA-CNN~\cite{fu2017look} is a reinforced attention proposal network to obtain discriminating attention regions and region-based feature representation of multiple scales. MA-CNN~\cite{zheng2017learning} was composed by three sub network to realise convolution, channel grouping and part classification respectively, which generates multiple object parts by clustering channels of feature maps into different groups. NTS~\cite{yang2018learning} enables a navigator agent as the region proposal network to detect multiple informative regions under the guidance from a teacher agent. PMG~\cite{du2020fine} adopts a progressive training strategy that fuses multi-granularity features.

The most relevant work to ours is MA-CNN~\cite{zheng2017learning}, which also highlights channel correlations and clusters them into groups. However, the setting in MA-CNN restricts itself to the last layer of the feature extractor, which ignores the low-level information in FGVC task. Besides, it adjusts the original network structure by setting a fixed number of individual part classifiers, which poses challenges to implement on other methods or tasks.

Compared with MA-CNN, the advantages of our work are two-folds. First, the proposed CDB is more flexible to be applied to any convolutional feature maps of classification model, and hence can make fully use of both high-level semantics and low-level details. Second, no adjustment to the original structure and no additional parameter/computation are involved during inference, which makes the method flexible to be integrated into existing networks.

\begin{figure*}[!t]
\begin{center}
   \includegraphics[width=1\linewidth]{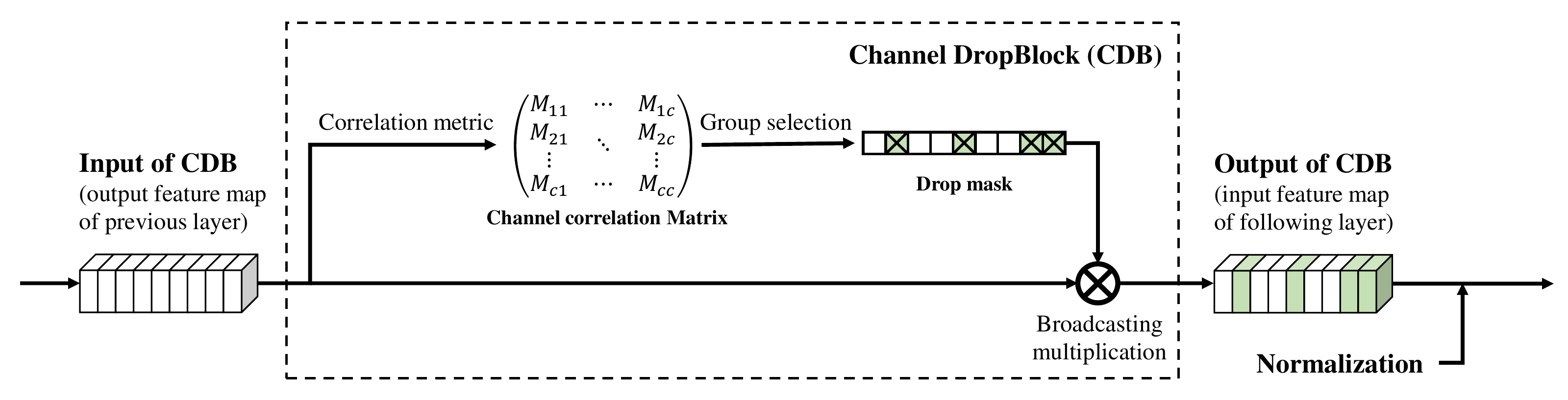}
\end{center}
   \caption{Illustration of the CDB block. The channel correlation matrix is generated based on different metrics. Then one channel and its corresponding visual group is randomly dropped by applying the drop mask to the input feature map, with dropped elements as 0, and 1 for otherwise.}
\label{fig:cdb_block}
\end{figure*}

\section{CDB: Channel DropBlock}\label{sec:Approach}
In this section, we present details of the proposed Channel DropBlock (CDB). It is a dropout-based regularization technique that can be easily applied to convolutional feature maps of classification model to improve feature representations. 
We first describe the motivation combined with comparison with relevant methods (Section \ref{subsec:Motivation}). We then describe the Channel DropBlock algorithm, which drops correlated channel groups based on channel correlation matrix (Section \ref{subsec:cdb alg} \& Section \ref{subsec:correlation metric}).

\subsection{Motivation}\label{subsec:Motivation}
As shown in previous work~\cite{zheng2017learning}, each channel of the convolutional features corresponds to a visual pattern. However, only parts of patterns contribute to the final prediction due to the co-adaptations between them, which will reduce inference accuracy especially when sub-categories are close and hard to distinguish (\emph{e.g.}, in FGVC tasks). While dropout~\cite{srivastava2014dropout} is effective to destruct co-adaptations in features,  it is less effective for convolutional feature channels since such channels are pairwise correlated, and the pattern about the input can still be sent to the next layer if we drop channels individually.
This intuition suggests us to mask out a correlated group of channels instead of a single channel to encourage the model to learn more discriminative parts.
The main motivation for CDB is to destruct the co-adaptations and induces the model to make full use of more discriminative features. This is achieved by randomly masking out a whole correlated channel group which only contributes to one visual evidence for the final prediction.

We initially developed CDB as an attention-based approach that specifically removes the most important channel groups from the input feature. This trail is similar to the idea in ADL~\cite{choe2019attention}, in that we develop an importance branch and a dropout branch, which are selected stochastically and work adversarially to highlight important channels and remove maximally activated group. We conduct this experiment as ablation studies, with limited improvements compared with randomly select one, because the random one can give more occlusion combinations, and are more likely to destruct co-adaptions between channels. We focus on Channel DropBlock with random selection for all of our experiments. 

Compared with MA-CNN~\cite{zheng2017learning} that clusters channels on the final feature map and settles individual classifier for each cluster, the proposed CDB is designed as a regularization block which is more flexible to applied on any convolutional feature maps of classification model.

Compared with SpatialDropout~\cite{tompson2015efficient}, CDB emphasizes that channels are correlated with each other, visual evidence can still be sent to the next layer with individually dropout.

Compared with DropBlock~\cite{ghiasi2018dropblock} that drops correlated units spatially, the proposed CDB calculates correlations channel-wise and can captures more precise visual evidences with two unique correlation metrics we provide.

\subsection{Channel DropBlock Algorithm}\label{subsec:cdb alg}
Algorithm~\ref{alg:CDB} and Figure \ref{fig:cdb_block} show the main procedure of the Channel DropBlock. Specifically, the input of CDB is a convolutional feature map $\mathbf{F} \in \mathbb{R}^{C\times H\times W}$, where $C$ is the number of channels, $H$ and $W$  denote to the height and width of $\mathbf{F}$, respectively. We obtain the correlation matrix $M \in \mathbb{R}^{C\times C}$ by calculating pairwise similarities between each feature channel (described in Section~\ref{subsec:correlation metric}). To obtain the drop mask, CDB first randomly selects one line in $M$, and produces the drop mask $M_d \in \mathbb{R}^C$ by setting top $\gamma$ most correlated elements as 0 and other elements as 1. The drop mask is then applied to the input feature map by broadcasting multiplication. In this way, features in a contiguous group are dropped together, which hides one certain discriminative pattern and encourages the model to learn other discriminative information that can also contribute to the final prediction. Similar to dropout, the proposed CDB only works in the training stage with normalization, no additional parameters and calculation costs are involved in the inference time. 


\begin{algorithm}[t]
\caption{ Training of Channel DropBlock.}  
\label{alg:CDB}  
\begin{algorithmic}[1]
  \Require
      Input feature map $\mathbf{F}$; 
      Drop rate $\gamma$;

    \State Calculate correlation matrix $M$
    \State Randomly select one channel in $M$ with equal probabilities and generate drop mask $M_d$ with top $\gamma$ most correlated channels setting as zero;
    \State Apply the mask: $\mathbf{F}=\mathbf{F}\times M_d$
    \State Normalize the features: $\mathbf{F}=\frac{1}{1-\gamma}\mathbf{F}$
    \State
    \Return $\mathbf{F}$;
\end{algorithmic}
\end{algorithm}

CDB has two main hyperparameters: $insert\_pos$ and $\gamma$. The parameter $insert\_pos$ indicates where the CDB is applied, and $\gamma$ controls the number of channels in the dropped group. 

\noindent\textbf{Influence of $insert\_pos$.}  As the structure of CNN getting deeper, the neurons in high layers are strongly respond to entire images and rich in semantics, but inevitably lose detailed information from small discriminative regions. With different setting of $insert\_pos$, the information of the input feature map differs. In our experiments, we settle an ablation study (described in Table~\ref{tab:insert_pos}) applying the proposed CDB block on variant layers in CNN.


\noindent\textbf{Setting the value of $\gamma$.} Another hyperparameter involves when we aggregate correlated channels into groups. Here we define $\gamma$ as the percentage of channels in a dropped group when conducting CDB. In practice, different correlation metrics will result in different cluster numbers and the number of channels in each cluster, so the setting of $\gamma$ distinct from the correlation metrics we choose.

\subsection{Channel Correlation}\label{subsec:correlation metric}
Ideally, a correlation metric should be symmetric and can cluster feature channels into different visual pattern groups. In this paper we examine two candidate metrics to evaluate the correlation between channels.

\begin{figure}[!t]
\begin{center}
  \includegraphics[width=1\linewidth]{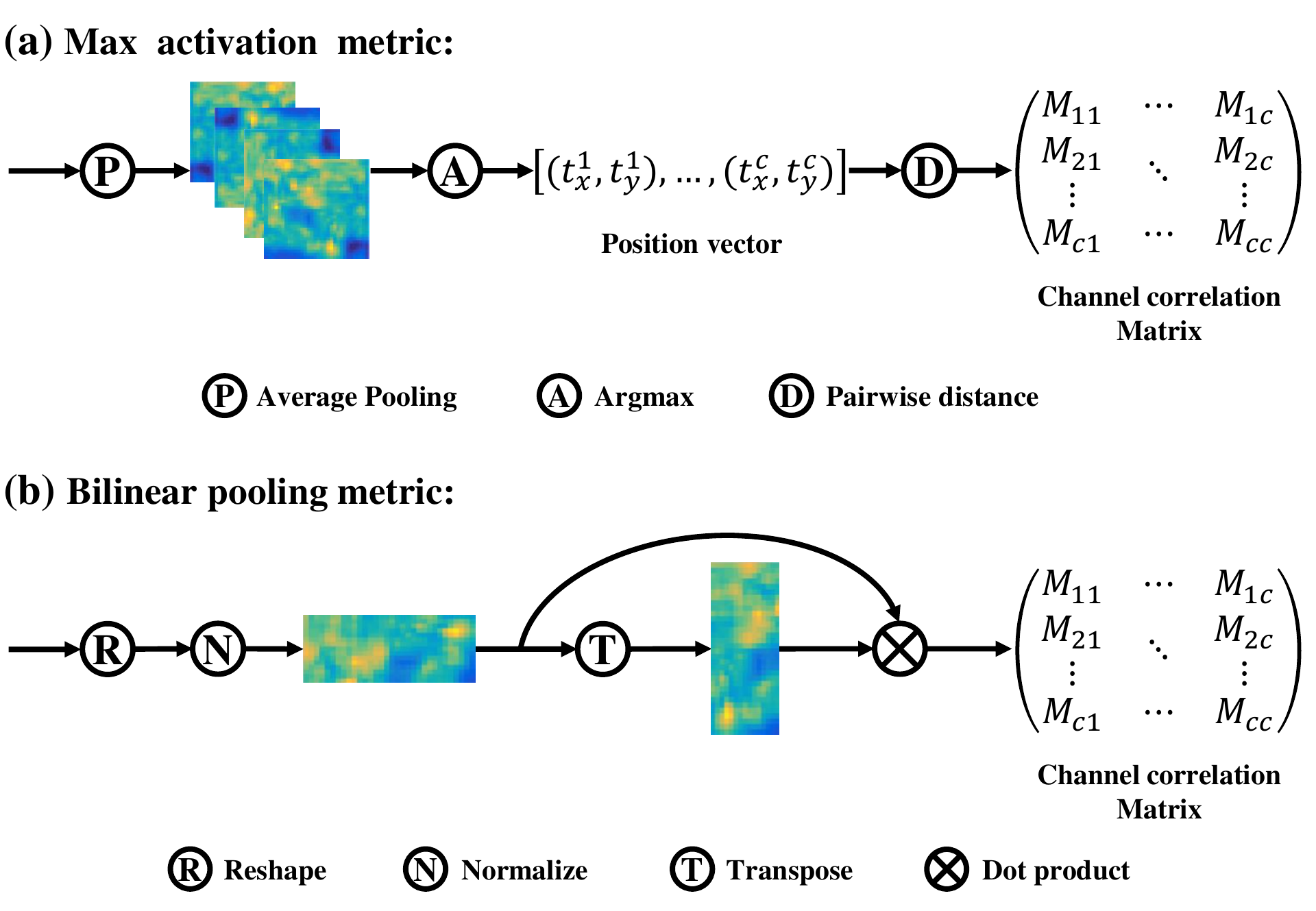}
\end{center}
  \caption{An illustration of channel correlation metrics: (a) max activation that groups channels into discriminative local region, and (b) bilinear pooling metric that groups channels based on visual pattern.}
\label{fig:metric_all}
\end{figure}

\noindent\textbf{Max activation metric.}
In order to distribute feature channels into groups, an intuitive idea is to divide them into different focused local regions. Inspired by the idea of MA-CNN \cite{zheng2017learning}, we treat channels with close maximal activation position as one pattern group. We conduct $3\times3$ average pooling to smooth feature maps and use the $argmax(\cdot)$ operation to get the coordinates of the peak response in each feature channel, which turns the input feature map $\mathbf{F}$ into a position matrix $P\in \mathbb{R}^{C\times 2}$ given by:
\begin{equation}
   P = [(t^1_x, t^1_y), (t^2_x, t^2_y), \cdot\cdot\cdot, (t^C_x, t^C_y)],
   \label{eq:position vection}
\end{equation}
where $t^i_x, t^i_y$ are the coordinates of the peak response of the $i^{th}$ channel. We then compute pairwise Euclidean distance between each activation position and obtain the correlation matrix $M$:
\begin{equation}
   M(i,j) = \Vert{(t_x^i,t_y^i)-(t_x^j,t_y^j)}\Vert^2.
   \label{eq:activation metric}
\end{equation}
In this metric, feature channels are grouped into discriminative local regions. Further more, it is a parameter-free metric that no learnable parameters are involved. Figure \ref{fig:metric_all} (a) shows the procedure of the max activation metric.

\noindent\textbf{Bilinear pooling metric.}
We also examine a correlation metric based on the bilinear pooling operator \cite{lin2015bilinear}, which calculates normalized cosine distance to measure channel similarities. In this approach, the input feature map $\mathbf{F}$ is reshaped into a matrix with a shape of $C\times HW$, which is denoted as $\mathbf{X}\in \mathbb{R}^{C\times HW}$. Then the reshaped matrix is fed through a normalization function followed by a bilinear pooling operator to get spatial relationship among channels:
\begin{equation}
   M \leftarrow \mathcal{N}(\mathbf{X})\mathcal{N}(\mathbf{X^T}),
   \label{eq:bp metric}
\end{equation}
where $\mathcal{N}(\cdot)$ indicates the L2 normalization function over the second dimension of the matrix. $\mathbf{X}\mathbf{X^T}$ is the homogeneous bilinear feature. Compared to the max activation metric, each channel group in this metric indicates one specific visual pattern. Similarly, no trainable parameters are involved in both the training phase and the inference phase. Figure \ref{fig:metric_all} (b) shows the procedure of the bilinear pooling metric.

\begin{table*}[t]
  \caption{Comparison results on CUB-200-2011, Stanford Cars, and FGVC-Aircraft datasets. CDB is applied on features from $conv2$ and $conv3$, with $\gamma$ set to $20\%$ for the max activation metric and $5\%$ for the bilinear pooling metric. CDB-MA and CDB-BP indicate CDB with the max activation metric and the bilinear pooling metric, respectively. The best and second-best results are marked respectively in bold and underlined fonts.}
  \begin{center}
    \setlength{\tabcolsep}{3mm}{
    \begin{tabular}{lcccc}
    \toprule
    Method & Backbone & CUB (\%) & CAR (\%) & AIR (\%) \\
    \midrule
    FT VGGNet~\cite{simonyan2014very} & VGG19 & 85.8 & 84.6 & 85.8 \\
    FT ResNet~\cite{he2016deep} & ResNet50 & 86.6 & 92.3  & 90.8 \\
    BCNN~\cite{lin2015bilinear} & VGG16 & 84.1 & 91.3 & 89.0 \\
    MA-CNN~\cite{zheng2017learning} & VGG19 & 86.5 & 92.8 & 89.9 \\
    MC-Loss~\cite{chang2020devil} & ResNet50 & 87.3 & 93.7 & 92.6 \\
    CIN~\cite{gao2020channel} & ResNet50 & 87.5 & 94.1 & 92.8 \\
    DCL~\cite{chen2019destruction} & ResNet50 & 87.8 & 94.5 & 93.0 \\
    PMG~\cite{du2020fine} & ResNet50 & 89.6 & \underline{95.1} & 93.4 \\
    \midrule
    CDB-MA (VGGNet) & VGG19 & 86.2 & 87.0 & 87.8 \\
    CDB-BP (VGGNet) & VGG19 & 86.1 & 86.6 & 85.9 \\
    CDB-MA (ResNet) & ResNet50 & 86.9 & 93.5 & 91.9 \\
    CDB-BP (ResNet) & ResNet50 & 87.2 & 93.2 & 91.5 \\
    \midrule
    CDB-MA (BCNN) & VGG16 & 84.6 & 91.9 & 90.4 \\
    CDB-BP (BCNN) & VGG16 & 84.5 & 91.7 & 90.0 \\
    CDB-MA (PMG) & ResNet50 & \textbf{89.9} & \textbf{95.4} & \underline{93.7} \\
    CDB-BP (PMG) & ResNet50 & \underline{89.7} & \textbf{95.4} & \textbf{93.8} \\
    \bottomrule
    \end{tabular}}%
  \end{center}
  \label{tab:main}%
\end{table*}%

\section{Experimental Results and Discussions}\label{sec:Experiments}

\textbf{Datasets.}
We evaluate the performance of the proposed method on three fine-grained benchmark datasets: CUB-200-2011 (CUB) \cite{wah2011caltech}, Stanford-Cars (CAR) \cite{krause20133d}, and FGVC-Aircraft (AIR) \cite{maji2013fine}, together with two classical image classification benchmark datasets: CIFAR-10 (C10) \cite{krizhevsky2009learning} and CIFAR-100 (C100) \cite{krizhevsky2009learning}.


\noindent\textbf{Implementation details.}
We use VGG19 \cite{simonyan2014very} and ResNet50 \cite{he2016deep} as backbone networks, and replace the origin classifier with an average pooling layer followed by a fully connected layer. We apply DCB with $\gamma$ set to $20\%$ for the max activation metric and $5\%$ for the bilinear pooling metric.
We use open-source Pytorch as our code-base, and train all the models on a single GTX 1080Ti GPU. Optimization is performed using Stochastic Gradient Descent with momentum of 0.9 and weight decay of 5e-4. 

For the FGVC datasets, the input images are resized to $448\times 448$. All of the models are pretrained on ImageNet and fine tuned for another 100 epochs with batches of 16 images. The initial learning rate is set to 0.001 and drops to 0 using cosine anneal schedule~\cite{loshchilov2016sgdr}.
For the C10/C100 datasets, the input images are first zero-padded with 4 pixels on each side, then randomly cropped into $32\times 32$. Models are trained from scratch for 200 epochs with each batch containing 128 images. The initial learning rate is set to 0.1 and drops to 0 using cosine anneal schedule.

\subsection{Fine-grained Image Classification}
The comparisons of our method with other state-of-the-art methods on three benchmark FGVC datasets are presented in Table \ref{tab:main}. The first block lists recent works for weakly supervised FGVC.

\noindent\textbf{Apply to baseline networks.}
The second block in Table \ref{tab:main} summarizes the performance of CDB on two baseline networks and three benchmark FGVC datasets. For the CUB dataset we follow the enhanced network setting and augmentation strategy in PMG \cite{du2020fine}, in which the classifier is combined with two fully connected layers, the input images are resize to a size of $550\times 550$, and randomly cropped to $448\times 448$ in the training time, centrally cropped to $448\times 448$ in the inference time.

We find that CDB with two correlation metrics outperforms the baselines on all FGVC datasets by a clear margin. Since FGVC focuses on differentiating sub-categories that share close similarities, it requires much more fine-grained visual evidence for prediction. The CDB block drops one entire visual group each iteration that induces model to distil other discriminative parts, thus being extremely suitable for this task.

\noindent\textbf{Apply to SOTA methods.}
The third block in Table \ref{tab:main} shows the results applying CDB on existing FGVC methods. We choose BCNN and PMG as benchmark models, the former one is a classical feature encoding-based approach which encode higher order information on features, the latter one is a SOTA approach which adopt a progressive training strategy that fuses multi-granularity features. CDB further improves the accuracies for a relative margin of $0.5\%$, $0.6\%$ ,$1.4\%$ on BCNN, and $0.3\%$, $0.3\%$, $0.4\%$ on PMG. 

\begin{figure*}[!t]
\begin{center}
   \includegraphics[width=0.9\linewidth]{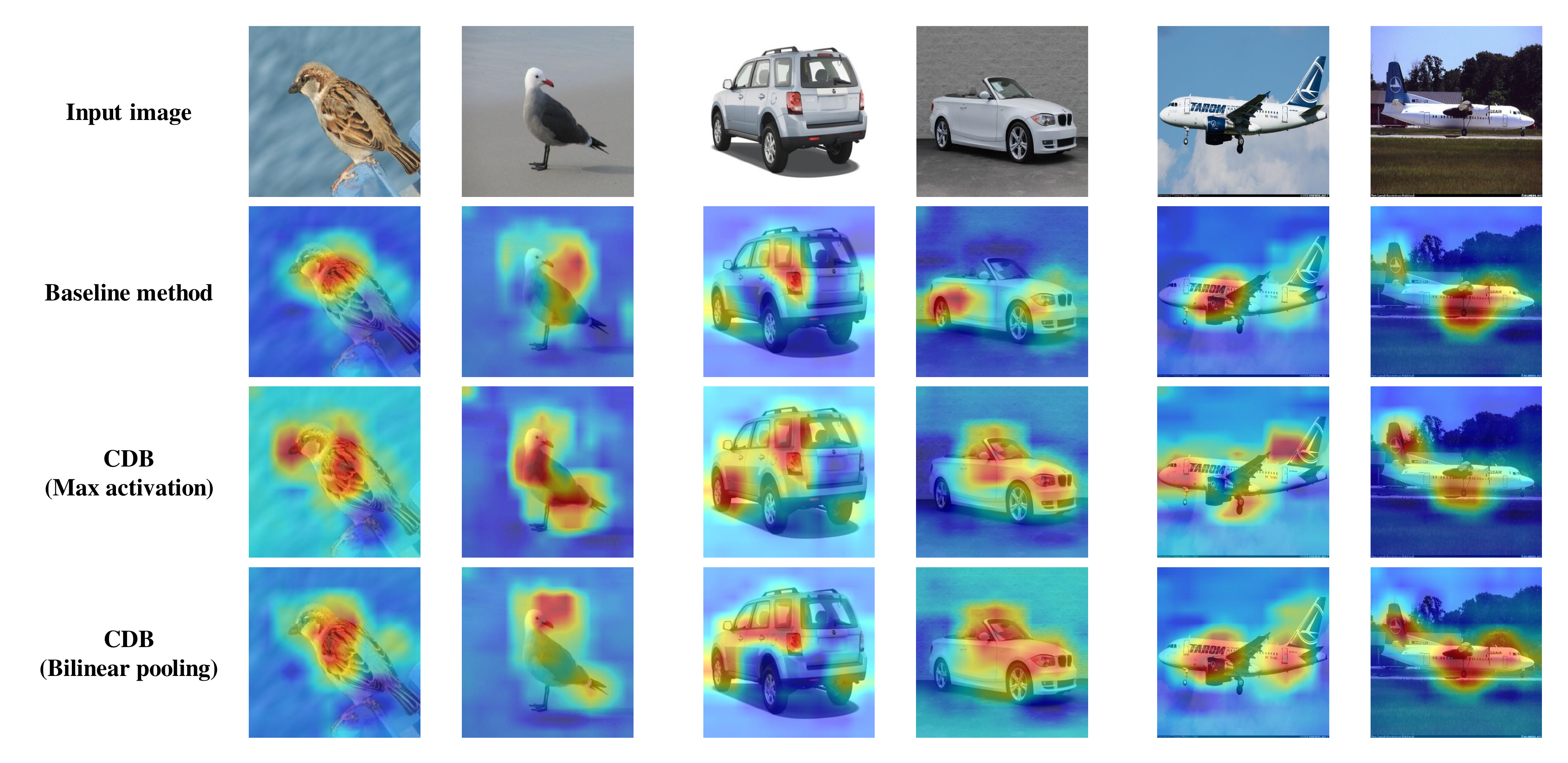}
\end{center}
   \caption{Visualization of the Gradient-weighted Class Activation Map (Grad-CAM) on FGVC samples with ResNet50 as the backbone model. The model trained with CDB tends to focus on multiple discriminative patterns. 
   }
\label{fig:visualization}
\end{figure*}

\subsection{Ablation study}
We conduct ablation studies to analyze the influence of insert position and dropout guidance, and compare CDB with other regularization techniques. We also settle experiments analyzing the performance of CDB on traditional image classification tasks. The following experiments are all conducted on the Stanford-Cars dataset with ResNet50 as backbone if not particularly mentioned.

\noindent\textbf{Where to apply CDB.}
In order to judge the influence of $insert\_pos$, we apply CDB on variant layers in CNN. Specifically, we define v1 to v5 indicating the last layer of $conv1$ to $conv5$ in ResNet50, and $pool1$ to $pool5$ in VGG19, and apply CDB with different $insert\_pos$ alone or their combinations. The experimental results in Table~\ref{tab:insert_pos} suggest that with the $insert\_pos$ on both v2\&v3 gives the best results. This is because the middle features in v2 and v3 contains both high-level semantics and low-level details, which contributes to the FGVC task.

\begin{table}[t]
  \centering
  \caption{Ablation study on insert position.}
    \begin{tabular}{lcccccc}
    \toprule
    \multirow{2}*{Metric} & \multicolumn{6}{c}{Accuracy (\%)} \\
    \cline{2-7}       
          & v1    & v2    & v3    & v4    & v5    & v2\&v3 \\
    \midrule
    MA & 92.9  & 93.2  & 93.2  & 92.6  & 93.0    & \textbf{93.5} \\
    \midrule
    BP & 93.1  & 92.7  & 93.0    & 92.8  & 92.9  & \textbf{93.2} \\
    \bottomrule
    \end{tabular}%
  \label{tab:insert_pos}%
\end{table}%

\begin{table}[t]
  \centering
  \caption{Comparison results with different drop guidance.}
    \begin{tabular}{lcc}
    \toprule
    Method & Random (\%) & Attention (\%) \\
    \midrule
    Baseline & 92.3  & 92.3 \\
    CDB-MA & \textbf{93.5}  & 92.6 \\
    CDB-BP & \textbf{93.2}  & 91.8 \\
    \bottomrule
    \end{tabular}%
  \label{tab:att-guided}%
\end{table}%

\begin{table}[t]
  \centering
  \caption{Comparison results with different regularization techniques.}
    \begin{tabular}{lccc}
    \toprule
    Method & CUB (\%) & CAR (\%) & AIR (\%) \\
    \midrule
    Baseline & 86.6 & 92.3 & 90.9 \\
    Dropout & 86.9 & 92.7 & 91.7 \\
    SpatialDropout & 86.9 & 92.8 & 91.5 \\
    Cutout & 87.1 & 92.7 & 90.9 \\
    DropBlock & 86.8 & 93.1 & 90.9 \\
    \midrule
    CDB-MA & 86.9 & \textbf{93.5} & \textbf{91.9} \\
    CDB-BP & \textbf{87.2} & 93.2 & 91.5 \\
    \bottomrule
    \end{tabular}%
  \label{tab:regularization techniques}%
\end{table}%


\begin{table}[t]
  \centering
  \caption{Comparison results on CIFAR-10 and CIFAR-100 datasets.}
    \begin{tabular}{lcc}
    \toprule
    Method & C10 (\%) & C100 (\%) \\
    \midrule
    ResNet50 & 95.1  & 78.1 \\
    ResNet50 + CDB-MA & 95.4  & \textbf{79.4} \\
    ResNet50 + CDB-BP & \textbf{95.5}  & 78.5 \\
    \midrule
    VGG19 & 93.7  & 71.8 \\
    VGG19 + CDB-MA & 93.7  & 72.5 \\
    VGG19 + CDB-BP & \textbf{93.9}  & \textbf{73.4} \\
    \bottomrule
    \end{tabular}%
  \label{tab:tradition_classification}%
\end{table}%

\noindent\textbf{Effect of dropout guidance.} We conduct experiments with different guidance to select the dropped channel group. Specifically, for randomly guidance we select one channel with equal probability and drop its corresponding correlated group. For the attention-guided one we follow the setting in ADL by developing an importance branch to learn channel attentions, and a dropout branch to drop the most important channel and its corresponding groups. The results in Table~\ref{tab:att-guided} suggest that randomly guided CDB can give more occlusion combinations to destruct co-adaptions between channels, and achieve significant performance improvements.

\noindent\textbf{Comparison with other regularization techniques.}
We compare the proposed CDB with different dropout based regularization techniques including dropout~\cite{srivastava2014dropout}, SpatialDropout~\cite{tompson2015efficient}, cutout~\cite{devries2017improved}, and DropBlock~\cite{ghiasi2018dropblock}. We conduct experiments on the FGVC datasets with ResNet50 as backbone, and train the model with different settings and report the best results. As shown in Table~\ref{tab:regularization techniques}, CDB 
achieves the best accuracy of $87.2\%$, $93.5\%$, and $91.9\%$ on three FGVC benchmark datasets, respectively, which confirms its significance.

\noindent\textbf{Traditional Image Classification.}
We conduct identical experiments on CIFAR-10 and CIFAR-100 dataset to evaluate the performance of CDB on traditional image classification task. CIFAR-10 has 10 distinct classes, such as cat, dog, car, and boat. CIFAR-100 contains 100 classes, but requires much more fine-grained recognition compared to CIFAR-10 as some classes are very visually similar. 
For example, it contains five different classes of trees: maple, oak, palm, pine, and willow. 

Table \ref{tab:tradition_classification} summarize the performance on traditional image classification. It can be observed that the improvement in C10 is limited while in C100 significant. As C100 requires much more fine-grained recognition compared to C10, it demonstrate that CDB is more effective in fine-grained tasks.

\subsection{Visualization}
In order to draw insights of the proposed method, we apply Grad-CAM~\cite{selvaraju2017grad} to visualize class activations of fine-grained samples on ResNet50 trained with and without CDB. 
We select test images from CUB-200-2011, Stanford-Cars, and FGVC-Aircraft, respectively. The second row of Figure~\ref{fig:visualization} shows the class activations of the baseline model and the third and fourth rows are visualization of the proposed method with two proposed correlation metrics. Consistent observations demonstrate that models trained with CDB not only dilate activations of the baseline model, but also distil other visual patterns to improve feature representations.

\section{Conclusion}
\label{sec:Conclusion}
In this paper we introduce a novel regularization technique, Channel DropBlock (CDB), which destructs feature channels from co-adaptations by clustering channels with correlation metrics and dropping a correlated channel group randomly during training. We demonstrate that CDB is more lightweight and effective to enhance feature representations and distil multiple discriminative patterns than existing FGVC methods. We conduct experiments on three widely tested fine-grained datasets, which confirm the superiority of the proposed method. Two particularly interesting directions for future work include exploring the method that group channels with adaptive size, and measuring channel correlations with integrated metrics.

\bibliographystyle{named}
\bibliography{ijcai21}

\end{document}